\documentclass[letterpaper]{article} 
\usepackage{aaai25}
\usepackage{times}  
\usepackage{helvet}  
\usepackage{courier}  
\usepackage[hyphens]{url}  
\usepackage{graphicx} 
\usepackage{natbib}  
\usepackage{caption} 
\usepackage{inconsolata}
\usepackage{makecell}
\usepackage{float}
\usepackage{bm}
\usepackage{booktabs}
\usepackage{multirow}
\usepackage{xcolor}
\usepackage{colortbl}
\usepackage{xspace}
\usepackage{subfig}
\usepackage{graphicx} 
\usepackage{color}
\usepackage{paralist}
\usepackage{acronym}
\usepackage{tabularx}
\usepackage[ruled,linesnumbered]{algorithm2e}
\usepackage{amsmath,amsfonts,amsthm}
\theoremstyle{definition}

\acrodef{LLM}{large language model}
\acrodef{NLP}{natural language processing}
\acrodef{KELE}{Knowledge Erasure for Large Language Model Editing}

\newcommand{\themodel}{\ac{KELE}\xspace}

\usepackage{xcolor}

\usepackage{newfloat}
\usepackage{listings}
\DeclareCaptionStyle{ruled}{labelfont=normalfont,labelsep=colon,strut=off} 
\floatstyle{ruled}
\newfloat{listing}{tb}{lst}{}
\floatname{listing}{Listing}
\nocopyright 

\title{Enhancing Multi-hop Reasoning through Knowledge Erasure \\in Large Language Model Editing}
\author{
    Mengqi Zhang\textsuperscript{\rm 1}\equalcontrib ,
    Bowen Fang\textsuperscript{\rm 2}\equalcontrib,
    Qiang Liu\textsuperscript{\rm 2},\\ 
    Pengjie Ren\textsuperscript{\rm 1},
    Shu Wu\textsuperscript{\rm 2},
    Zhumin Chen\textsuperscript{\rm 1},
    Liang Wang\textsuperscript{\rm 2}
}
\affiliations{
    \textsuperscript{\rm 1}School of Computer Science and Technology, Shandong University\\

\textsuperscript{\rm 2}New Laboratory of Pattern Recognition (NLPR) \\
  State Key Laboratory of Multimodal Artificial Intelligence Systems  (MAIS)\\
  Institute of Automation, Chinese Academy of Sciences \\
        \textrm{\{mengqi.zhang, renpengjie, chenzhumin\}@sdu.edu.cn, bwn.fang@gmail.com\\ {\{qiang.liu,shu.wu,wangliang\}@nlpr.ia.ac.cn}}
}

\usepackage{bibentry}

\begin{document}

\maketitle
\definecolor{dgreen}{RGB}{0, 176, 80}
\begin{abstract}
\Acp{LLM} face challenges with internal knowledge inaccuracies and outdated information. Knowledge editing has emerged as a pivotal approach to mitigate these issues. Although current knowledge editing techniques exhibit promising performance in single-hop reasoning tasks, they show limitations when applied to multi-hop reasoning.  Drawing on cognitive neuroscience and the operational mechanisms of \acp{LLM}, we hypothesize that the residual single-hop knowledge after editing causes edited models to revert to their original answers when processing multi-hop questions, thereby undermining their performance in multi-hop reasoning tasks. To validate this hypothesis, we conduct a series of experiments that empirically confirm our assumptions. Building on the validated hypothesis, we propose a novel knowledge editing method that incorporates a \underline{K}nowledge \underline{E}rasure mechanism for \underline{L}arge language model \underline{E}diting (KELE). Specifically, we design an erasure function for residual knowledge and an injection function for new knowledge. Through joint optimization, we derive the optimal recall vector, which is subsequently utilized within a rank-one editing framework to update the parameters of targeted model layers.
Extensive experiments on GPT-J and GPT-2 XL demonstrate that KELE substantially enhances the multi-hop reasoning capability of edited LLMs.
\end{abstract}

%

\section{Introduction}

Large Language Models (LLMs) have achieved significant success in a wide range of Natural Language Processing (NLP) tasks \cite{zhao2023survey}. However, the knowledge embedded within LLMs can sometimes be factually incorrect or outdated, which limits their overall effectiveness. To address these limitations, knowledge editing methods have been proposed, offering a more efficient and precise approach to updating the knowledge in LLMs. These methods have attracted considerable attention from researchers in recent years.

Knowledge editing can be broadly categorized into methods that preserve LLMs’ parameters and methods that modify them. The former approach mainly involves storing edited examples in an external knowledge base or adding additional knowledge parameters to adjust the model's output for specific queries, such as SERAC \cite{mitchell2022memory} and T-patcher \cite{huangtransformer}. The latter approach generally entails directly modifying the model's parameters to alter its output. This category includes meta-learning methods that employ a hyper-network to learn the weight changes for editing LLMs, such as KE \cite{de2021editing} and MEND \cite{mitchell2021fast}. It also encompasses fine-tuning-based methods, such as InstructEdit \cite{tian2024instructedit}, which perform editing through efficient parameter tuning. Additionally, locate-then-edit methods, such as ROME \cite{meng2022locating} and MEMIT \cite{meng2022mass}, specifically target and update parameters corresponding to particular pieces of knowledge.

\begin{figure}
  \centering
  \includegraphics[width=0.9\linewidth]{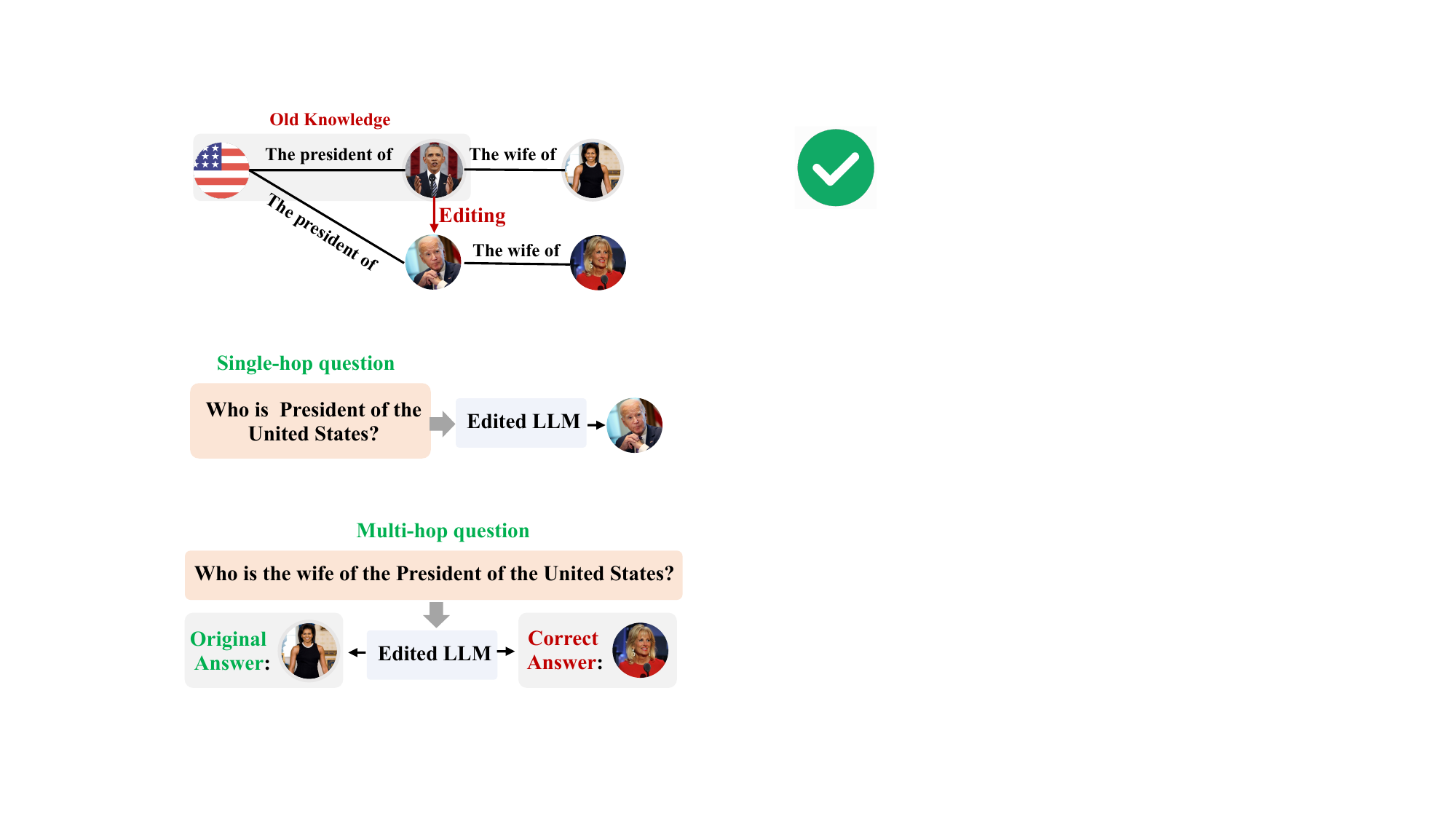}
  \caption{Example of Knowledge Editing }
  \label{mov1}
 \end{figure}

Although these editing methods have demonstrated promising results in single-hop reasoning evaluations, they still face significant challenges in multi-hop reasoning \cite{zhong2023maquake}. As illustrated in Figure \ref{mov1}, after editing the single-hop knowledge from “\emph{The President of the USA is Obama}” to “\emph{The President of the USA is Biden},” the edited model can easily answer the single-hop question, “\emph{Who is the President of the USA?}” However, it struggles with multi-hop questions, such as “\emph{Who is the wife of the President of the USA?}”

To better understand this challenge in knowledge editing for LLMs, we first analyze this problem from a cognitive neurological perspective. When the brain receives new information, it can activate neurons associated with related old memories, a phenomenon known as Memory Association \cite{roediger1995creating,schacter1998priming,kahana2012foundations}. This occurs because of the connectivity within neural networks, where the pathways of old memories are easily reactivated by relevant stimuli, thereby facilitating more efficient encoding and processing of new information. LLMs exhibit a similar mechanism, where related knowledge stored in their parameters is activated and integrated during reasoning \cite{geva2021transformer}.

\begin{figure}
  \centering
  \includegraphics[width=1\linewidth]{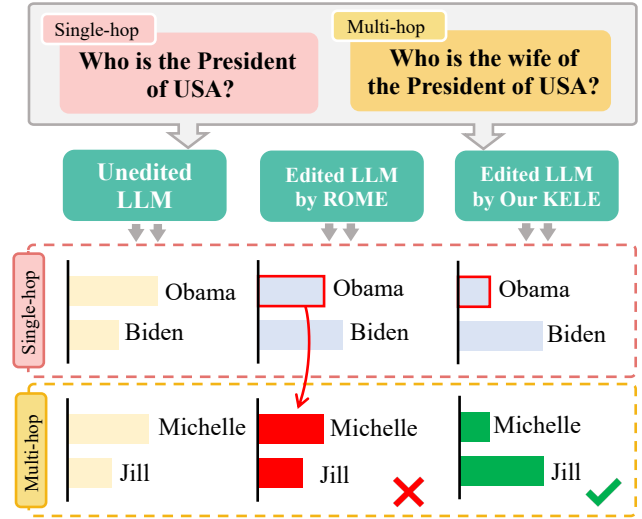}
  \caption{Single-hop and Multi-hop evaluation of Unedited LLM, LLM edited by ROME and our KELE. When confronted with a multi-hop question, the residual old single-hop knowledge (The President of the USA is Obama) in the LLMs edited by ROME prompts the model to generate the \textbf{original answer},  \textcolor{red}{\textbf{Michelle}} (Obama’s wife), instead of the \textbf{correct answer}, \textcolor{dgreen}{\textbf{Jill}} (Biden’s wife).}
  \label{mov2}
 \end{figure}

Building on these insights, we hypothesize the following reason for the poor performance of edited LLMs on multi-hop reasoning tasks: \textbf{\emph{LLMs retain a portion of single-hop old knowledge even after editing. When handling multi-hop questions related to the edited knowledge, the residual knowledge tends to prompt the models to produce original answers to these questions, thereby weakening their multi-hop reasoning ability.}} For example, if the single-hop knowledge in the LLM is edited from “\emph{The President of the USA is Obama}” to “\emph{The President of the USA is Biden},” a portion of old knowledge “\emph{The President of the USA is Obama}” may still be retained and reactivated within the model. As shown in Figure \ref{mov2}, when asked the multi-hop question “\emph{Who is the wife of the President of the USA?}”, the residual single-hop knowledge might cause the model to generate the original answer to the multi-hop question, \emph{Michelle (Obama’s wife)}, instead of the correct answer, \emph{Jill (Biden’s wife)}.

To verify this hypothesis, we investigate the relationship between the residual old knowledge in LLMs and their responses to multi-hop questions (Section \ref{ans-hop}). We define the Retain Score as a metric to quantify the residual old knowledge $(s,r,o)$ for each edit sample $(s, r, o, o*)$, utilizing the output logit score of $o$ under the prompt $p(s,r)$. As illustrated in Figure \ref{impact2}, the higher the residual old knowledge in the edited LLM, the more likely it is to provide the original answers to multi-hop questions, resulting in a lower proportion of correct answers. Therefore, erasing the residual old knowledge offers a promising insight for improving the performance of edited LLMs on multi-hop reasoning tasks.

Based on the this hypothesis, we propose a simple yet effective method for large language model editing, termed \themodel (Section \ref{method}). Specifically, within the rank-one editing framework, we develop an old knowledge erasure function and a new knowledge injection function to jointly optimize and obtain the recall vector. This approach eliminates the interference of old knowledge while injecting new knowledge. Finally, the model parameters are updated in a single step using the recall vector and the subject representation through the rank-one update formula.

We summarize our contributions as follows:
\begin{itemize}
\item We investigate and validate the impact of residual old single-hop knowledge in edited LLMs on multi-hop reasoning tasks, demonstrating that such residual knowledge may cause edited LLMs to revert to original answers when faced with multi-hop questions.
\item We integrate a knowledge erasure strategy into model editing and propose \themodel, a simple yet effective editing method to enhance the multi-hop reasoning performance of edited LLMs.
\item We conduct extensive experiments on GPT-2 XL and GPT-J, showing that KELE significantly enhances the multi-hop reasoning ability of edited models.
\end{itemize}

\section{Related Work}
In this section, we review related research on knowledge editing and its challenges in multi-hop reasoning.

\subsubsection{Methods for preserving models' parameters} These methods typically store edited examples in an external knowledge base and guide the LLMs' output for specific queries by retrieving relevant knowledge. For instance, SERAC \cite{mitchell2022memory} employs a gating network along with an auxiliary model specifically designed to handle edited knowledge. T-patcher \cite{huangtransformer} introduces extra trainable parameters in the last layer of the FFN to 
 correct LLMs. However, these methods face a critical scalability issue: the complexity of managing the external model increases with each new edit, which may limit their practical usability.  

\subsubsection{Methods for modifying models' parameters} These methods, including meta-learning, locat-and-edit, and fine-tuning-based approaches, edit LLMs by directly modifying their parameters. Meta-learning methods generate updated weights for LLMs by training a hyper-network. For example, KE \cite{de2021editing} uses a bi-directional LSTM to predict model weight updates, but it faced challenges with larger models due to their vast parameter spaces. To address this issue, MEND \cite{mitchell2021fast} employs a low-rank decomposition of fine-tuning gradients, providing an efficient mechanism for updating LLM weights. Locate-and-edit methods focus on identifying specific parameters associated with particular knowledge within \acp{LLM}, aiming for more interpretable and precise knowledge editing. Early efforts, such as KN \cite{dai2022knowledge}, introduce a knowledge attribution method to identify knowledge neurons but struggles to precisely modify the model’s weights. 
ROME \cite{meng2022locating} method employs causal tracing to identify knowledge-relevant layers and then edits the corresponding FFN module. MEMIT \cite{meng2022mass} further enhances this approach by improving the objective function and enabling multi-layer edits for batch editing. Recently, significant advancements in efficient parameter-tuning methods \cite{hulora,ren2024mini} for supervised fine-tuning of LLMs have led to the development of fine-tuning-based editing methods \cite{ni2023forgetting,gangadhar2024model}. For example, \cite{gangadhar2024model} utilizes LoRA \cite{hulora} and data augmentation strategies to directly fine-tune the LLMs, achieving the desired editing performance. F-Learning \cite{ni2023forgetting} proposes a new paradigm of knowledge updating during supervised fine-tuning, employing parametric arithmetic to forget old knowledge and learn new knowledge, thereby resolving contradictions between old and new knowledge. 

\subsubsection{Multi-hop reasoning in knowledge editing} In recent years, several studies have aimed to enhance the performance of edited LLMs in multi-hop reasoning tasks. \cite{zhong2023maquake} introduces the \textsc{MQuAKE} dataset, specifically designed to evaluate the multi-hop reasoning capabilities of edited LLMs. GLAME \cite{zhang2024knowledge} leverages external knowledge graphs to capture the impact of target knowledge changes on high-order knowledge within LLMs. However, its reliance on external information constrains its applicability in certain scenarios. Additionally, \cite{ju2024investigating} finds that LLMs often rely on factual shortcuts from pre-training corpora during reasoning, which contributes to the poor performance of edited models in multi-hop reasoning tasks. Unlike this study, we hypothesize that the retention of old knowledge triggers the generation of original answers in multi-hop questions, thereby weakening the performance of edited models in these tasks. We validate this hypothesis through a series of experiments and propose a knowledge-erasure-based editing strategy to mitigate this issue.

\section{Preliminaries}
In this section, we introduce the definition of knowledge editing and outline the corresponding tasks under single-hop and multi-hop evaluations.Additionally, we describe the rank-one model editing framework employed in our study.

\subsubsection{\textbf{Knowledge Editing for LLMs}}
Knowledge editing \cite{yao-etal-2023-editing} refers to the process of altering the behavior of an LLM $\mathcal{F}$'s to change encoded knowledge from $(s, r, o)$ to the new knowledge $(s, r, o^*)$. Here, knowledge is represented as a triple, with $s$ as the subject, $r$ as the relation, and $o$ as the object. Each editing instance $e$ is denoted as $(s,r,o,o^*)$, and the LLM after editing is referred to as $\mathcal{F}'$.

\subsubsection{\textbf{Single-hop Evaluation in Knowledge Editing}}
Single-hop evaluation assesses whether an edit $(s,r,o,o^*)$ is successful in an edited LLM $\mathcal{F}'$. This evaluation constructs prompts $p(s,r)$ based on the subject $s$ and relation $r$, and measures 
 the performance of $\mathcal{F}'$ using Efficacy, Paraphrase and Specificity metrics \cite{yao-etal-2023-editing}.

\subsubsection{\textbf{Multi-hop Evaluation in Knowledge Editing}}
Multi-hop evaluation examines whether the edited LLMs can effectively utilize the updated knowledge for reasoning in multi-hop tasks. Given a chain of facts $(s_1, r_1, o_1),..., (s_n, r_n, o_n)$, where the object of the $i$-th fact also serves as the subject of the next fact in the chain, i.e., $o_i=s_{i+1}$, a multi-hop question $p(s_1,r_1,..,r_n)$ can be constructed, with the answer being $o_n$. For example, with a chain consisting of two facts, \emph{(USA, president of, Obama)} and \emph{(Obama, wife of, Michelle)}, one can write a $2$-hop question: \emph{Who is the wife of the president of USA?}. Once one or more facts in the chain are edited, e.g., \emph{(USA, president of, Obama)} is edited to \emph{(USA, president of, Biden)}, the edited LLM must utilize the new knowledge to answer the multi-hop question. The model's response should change from the original answer \emph{Michelle} to the correct answer \emph{Jill}. 
%

\subsubsection{Rank-One Model Editing}
Rank-One Model Editing (ROME) \cite{meng2022locating} is a Locate-then-edit method that presupposes factual knowledge is stored within the Feedforward Neural Networks (FFNs), conceptualized as key-value memories \cite{geva2021transformer,kobayashi2023feed}. The output of the $l$-th layer FFN for the $i$-th token is given by:
    \begin{equation}
        \mathbf{v}_i^l = f(\mathbf{W}_{in}^l \cdot \mathbf{h}_i^{l-1}) \cdot \mathbf{W}^l,
    \label{fnn}
    \end{equation}
    where $f(\cdot)$ denotes the activation function, and $\mathbf{h}_i^{l-1}$ is the FFN input. For simplicity, the superscript $l$ is omitted in the following discussion. 
    
    In this context, $f(\mathbf{W}_{in} \cdot \mathbf{h}_i)$ functions as the keys, denoted as $\mathbf{k}_i$. The outputs of the subsequent layer represent the corresponding values. Utilizing casual tracing \cite{pearl2022direct,vig2020investigating}, this method identify a specific FFN layer for editing and updates the weight $\mathbf{W}$ of the second layer by solving a constrained least-squares problem:
\begin{align}
\begin{aligned}
& {\text{minimize}}
& & \|\mathbf{{W}}\mathbf{K}-\mathbf{V}\|, \\
& \text{subject to}
& & \mathbf{{W}}\mathbf{k}_* = \mathbf{v}_*.
\end{aligned} 
\end{align}
where the objective function aims to preserve the knowledge unrelated to the edited sample within the LLM. Here, $\mathbf{K} = [\mathbf{k}_1;\mathbf{k}_2;,\dots,;\mathbf{k}_p]$ denotes the sets of keys encoding subjects unrelated to the edited fact, and $\mathbf{V} = [\mathbf{v}_1;\mathbf{v}_2;,\dots,;\mathbf{v}_p]$ represents the corresponding values. The constraint ensures that the edited knowledge is incorporated into the FFN layer by enabling the key $\mathbf{k}_*$ (encoding subject $s$) to retrieve the value $\mathbf{v}_*$ about the new object $o^*$.

As explicated in \cite{meng2022locating}, a closed-form solution to the optimization problem can be derived: 
\begin{equation}
    \mathbf{\hat{W}} = \mathbf{W}+ \frac{(\mathbf{v}_*-\mathbf{W}\mathbf{k}_*)(\mathbf{C}^{-1}\mathbf{k}_*)^\mathrm{T}}{(\mathbf{C}^{-1}\mathbf{k}_*)^\mathrm{T} \mathbf{k}_*},
    \label{solution}
\end{equation} 
where $\mathbf{C}=\mathbf{K}\mathbf{K}^\mathrm{T}$ is a constant matrix, precomputed by estimating the uncentered covariance of $\mathbf{k}$ based on a sample of Wikipedia text (Appendix \ref{imp}). Thus, solving the optimal parameter $\mathbf{\hat{W}}$ is transformed into calculating subject representation $\mathbf{k}_*$ and recall vector $\mathbf{v}_*$.

\section{Analysis of the Impact of Old Knowledge on Multi-hop Reasoning}
\label{ans-hop}
In this section, we validate our hypothesis by examining the impact of old knowledge on the performance of edited LLMs in multi-hop reasoning. We select the representative multi-hop reasoning evaluation dataset, \textsc{MQuAKE} \cite{zhong2023maquake}, to conduct experiments. Each instance in \textsc{MQuAKE} is represented as $d=(\mathcal{E},\mathcal{Q}, a,a^*)$. Here, $\mathcal{E}$ denotes the set of single-hop edits $e=(s,r,o,o^*)$, $\mathcal{Q}$ represents multi-hop questions evaluating editing performance, and $a$ and $a^*$ are the original answer and correct answer to $\mathcal{Q}$. Further details of \textsc{MQuAKE} are provided in Section \ref{exper} and Appendix \ref{dataset}. 
 
\subsection{Retain Score}
We first define a metric to quantify the retention of old knowledge in the LLM. In cognitive neuroscience, memory activation is often measured by the intensity of neural activity. Analogously, in LLMs, the logit vector can serve as an indicator of the model’s memory activation strength. Building on this concept, we introduce the \textbf{Retain Score (RS)} indicator for each edit sample $e=(s, r, o, o^*)$ to measure the residual presence of the old knowledge $(s, r, o)$. 

When an LLM is given an input prompt, it generates the next token based on the logit vector produced by its final layer. A higher logit value for a token indicates greater model confidence in generating that token, corresponding to stronger memory activation. Consequently, we use the logit value as a measure of the model’s retention of old knowledge. To ensure a consistent assessment of retention across different editing instances, we standardize the logit vectors to eliminate variations from varying logit distributions:
\begin{equation}
    \mathbf{RS}(e) = \frac{D_o-\mu}{\sigma},
\end{equation} 
where $D$ represents the logit vector produced by the final layer of the LLM, $D_o$ is the logit score of $o$, while $\mu$ and $\sigma$ denote the mean and standard deviation of the logit vector $D$, respectively. 

\begin{figure}[t]
	\centering
	\subfloat[Unedited GPT-J]{\includegraphics[scale=0.33]{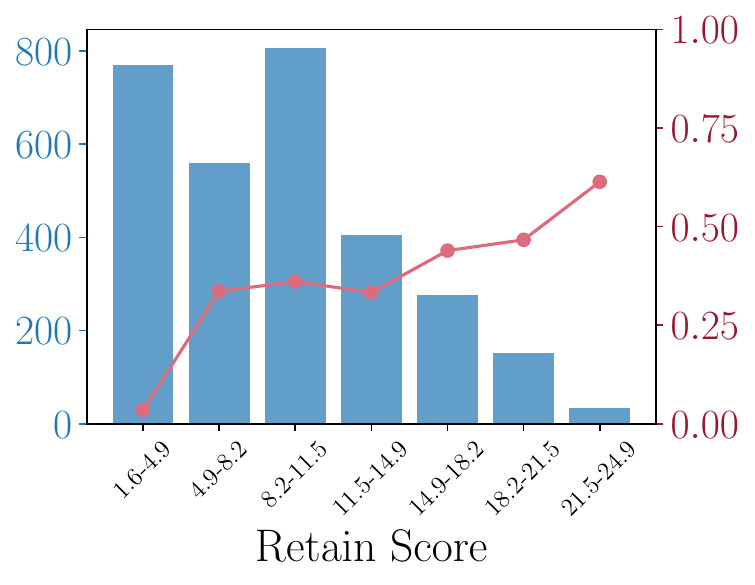}
 
 \label{impact1}
	}
	\subfloat[Edited GPT-J]{\includegraphics[scale=0.33]{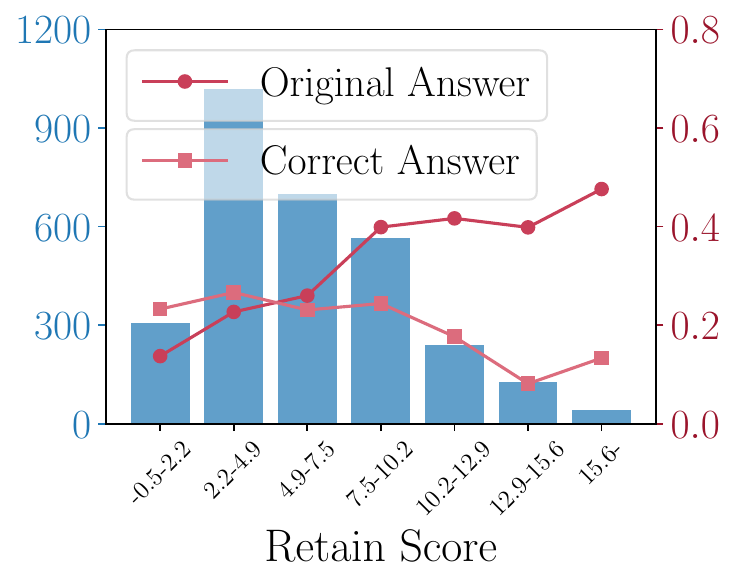}
  \label{impact2}
	}
	\caption{(a) The accuracy of single-hop answer generated by unedited GPT-J . (b)The accuracy of original and correct answers generated by edited GPT-J. The left y-axis represents the number of instances within each Retain Score interval, while the right y-axis indicates the accuracy.
	}
	\label{impact}
\end{figure}

\subsubsection{The reasonable of Retain Score} To validate the reasonableness of the Retain Score, we first divide the RS values of all $e$ in the dataset into different intervals. For each interval, we then calculate the probability that the unedited model correctly answers $o$ given the prompt $p(s,r)$. The experimental results, as shown in Figure \ref{impact1}, indicate that as the RS value increases, the accuracy of the unedited model’s responses also increases. This suggests that the model’s sensitivity to the corresponding knowledge strengthens as the RS value rises, demonstrating that the RS metric effectively measures the retention of old knowledge.

\subsection{Impact of Old Knowledge on Multi-hop Reasoning}
To further investigate the impact of residual old knowledge on multi-hop reasoning, we apply the ROME method to GPT-J and explore the relationship between the Retain Score and the accuracy of answering multi-hop questions. 

Specifically, for each instance $d$, we first calculate the accumulated old single-hop knowledge of all edit samples $\mathcal{E}$ in the edited models:  
\begin{equation}
    \mathbf{RS}(d) = \sum_{e\in \mathcal{E}}\mathbf{RS}(e).
\end{equation}
We then divide the dataset into different subsets based on the varying ranges of the Retain Score of the instances. For each subset, we calculate the accuracy of the edited model in answering the original and correct answers to the multi-hop questions. The results are shown in Figure \ref{impact2}.

As illustrated in Figure \ref{impact2}, we observe that as the Retain Score value increases, the edited models show a significant improvement in accuracy when providing the original answers to multi-hop questions. However, the accuracy of the edited model in providing correct answers decreases as the Retain Score rises. This suggests that as the amount of retained old knowledge increases, the model becomes more likely to favor the original answers, thereby diminishing its ability to generate correct responses to multi-hop questions.

These experiments validate that \textbf{\emph{LLMs retain traces of old single-hop knowledge after editing, which significantly motivates them to revert to original answers for multi-hop questions and undermines their performance in providing correct answers.}} Therefore, eliminating residual old knowledge during the editing process is crucial for enhancing the accuracy of LLMs in multi-hop reasoning.

\section{Methodology}
\label{method}
\begin{figure}
  \centering
  \includegraphics[width=1\linewidth]{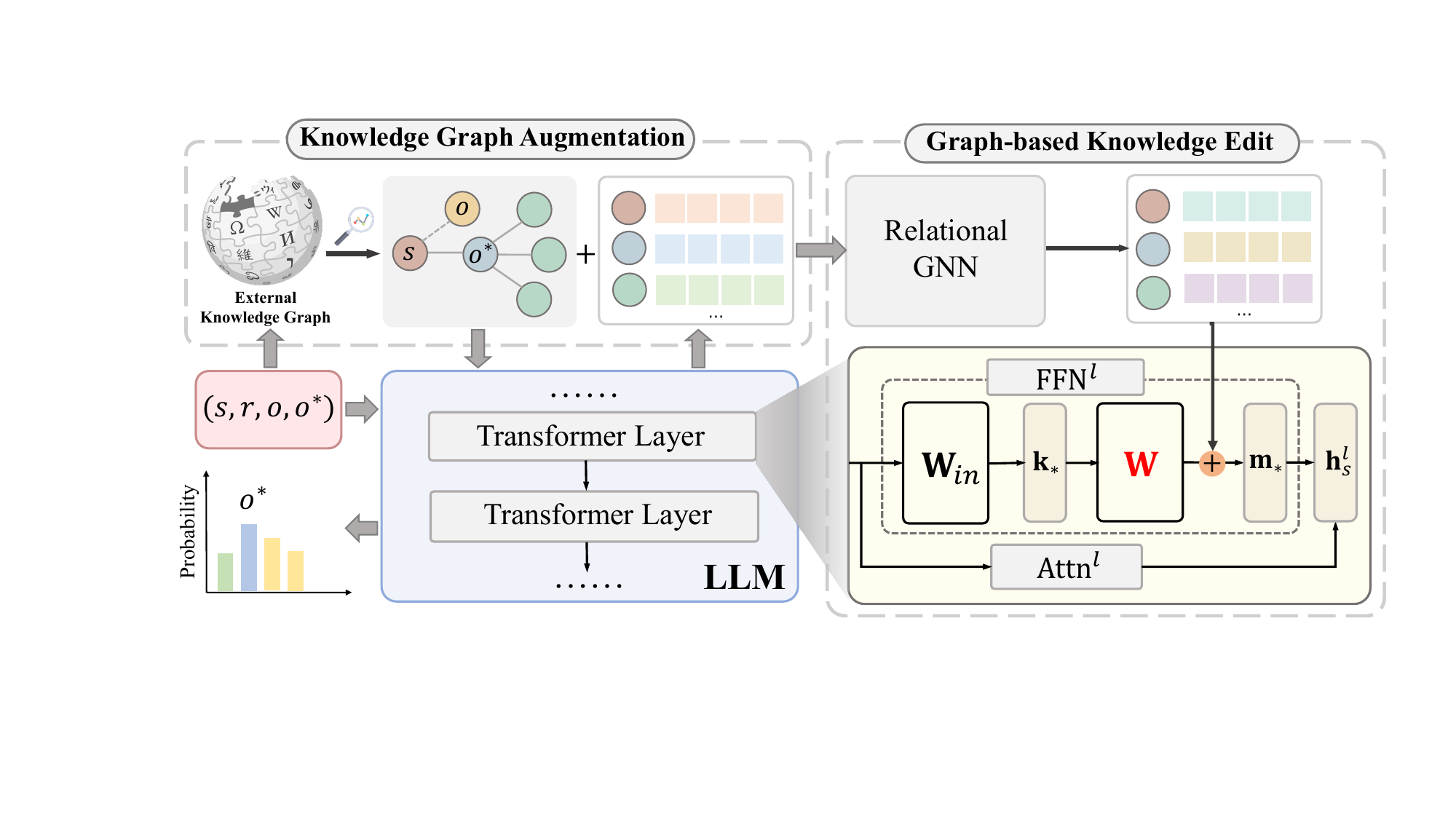}
  \caption{An illustration of KELE architecture. First, we use the old knowledge erasure function and the new knowledge injection function to derive the recall vector $\mathbf{v}_*$. Then, we compute the subject representation $\mathbf{k}_*$. Finally, the parameters are updated using the rank-one update formula.}
  \label{framework}
 \end{figure}

In this section, we introduce the proposed \themodel, with its architecture depicted in Figure \ref{framework}. The \themodel framework integrates a knowledge erasure strategy within the rank-one model editing framework \cite{meng2022locating}. Specifically, \themodel targets a specific layer $l$ and transforms knowledge editing into two key operations: old knowledge erasure and new knowledge injection, which together are used to compute the recall vector $\mathbf{v}_*$. Subsequently, $\mathbf{v}_*$, along with the subject representation $\mathbf{k}_*$, is applied in Equation (\ref{solution}) to update the parameters of the second layer of the FNN, thereby completing the knowledge editing process.

\subsection{Computing $\mathbf{v}_*$ to Recall New Knowledge}
To effectively edit new knowledge while minimizing the negative impact of old knowledge on multi-hop reasoning, we construct an old knowledge erasure function and a new knowledge inject function, which are jointly optimized to obtain $\mathbf{v}_*$. In this process, we optimize the learnable parameter vector $\mathbf{h}$ to modify the original value vector $\mathbf{v}_s^{l}$, resulting in the optimal vector $\mathbf{v}_*=\mathbf{v}_s^{l}+\mathbf{h}$. 

\subsubsection{\bf Old knowledge erasure function} To mitigate the influence of residual old knowledge that prompts the edited LLM to generate original answers in response to multi-hop question, we develop an old knowledge erasure function for calculating $\mathbf{v}_*$. Specifically, for each edit sample $(s,r,o, o^*)$, we introduce a max-margin loss aimed at reducing the likelihood that the edited LLM generates $o$ in response to the prompt $p(s,r)$. This is achieved by lowering the rank of $o$ in the logit vector of the final layer. The erasure function is defined as:
\begin{equation}
    \mathcal{L}_e = \max\limits_{{\mathcal{F}(\mathbf{v}_s^l+=\mathbf{h})}}(0, D_o-D_{k}),
    \label{eq-e}
\end{equation} 
where $D_o$ is the logit score of $o$ and $D_{k}$ is the $k$-th top value in logit vector $D$. Simply minimizing the probability of $o$ appearing can significantly reduce the retention of old knowledge $(s, r, o)$, but it may also cause substantial changes to the model, leading to negative impacts. Therefore, the margin loss uses $D_k$ to control the intensity of knowledge erasure, optimizing the logit value of $o$ only when its rank is whithin the top $k$. $\mathcal{F}(\mathbf{v}_s^{l}+=\textbf{h})$ indicates the LLM's inference alteration through the hidden state $\mathbf{v}_s^{l}$ modification to $\mathbf{v}_s^{l}+\mathbf{h}$.

\subsubsection{\bf New knowledge injection function} For each edit sample $(s,r,o,o^*)$, our second objective is to refine the parameter vector $\mathbf{h}$ enables the LLM to accurately predict the target object $o^*$. Accordingly, the knowledge injection loss function is defined as:
\begin{align}
    \mathcal{L}_p = -&\frac{1}{N} \sum^N_{j=1}\log\mathrm{P}_{\mathcal{F}(\mathbf{v}_s^{l}+=\mathbf{h})}[o^*\mid x_j \oplus p(s, r)],
\end{align}
where $x_j$ is the random prefix generated by the LLM to foster optimization robustness.

To mitigate the impact of above operations on the intrinsic of $s$ within the LLM, we minimize the KL divergence between $\mathcal{F}(\mathbf{v}_s^{l}+=\mathbf{h})$ and the original model $\mathcal{F}$ \cite{meng2022locating}:
\begin{equation}
    \mathcal{L}_a = D_{\text{KL}}\left(\mathrm{P}_{\mathcal{F}(\mathbf{v}_s^{l}+=\mathbf{h})}[x \mid p '] \parallel \mathrm{P}_{\mathcal{F}}[x \mid p’] \right),
\end{equation} 
where $p'$ denotes prompts in the form of "{subject} is a".

Ultimately, the parameter $\mathbf{h}$ is optimized by minimizing the following objective function:
\begin{align}
\label{loss}
    \mathcal{L} = \mathcal{L}_e + \mathcal{L}_p  + \lambda \mathcal{L}_a,
\end{align}
where $\lambda$ adjusts the regularization strength. Throughout the optimization process, the parameters of the LLM remain unchanged. 

\subsection{\bf Computing $\mathbf{k}_*$ to Represent Subject} 
For each edit sample $(s,r,o, o^*)$, the subject representation $\mathbf{k}_*$ is calculated by 
\begin{equation}
\label{k}
    \mathbf{k}_* = \frac{1}{N}\sum_{j=1}^{N}f(\mathbf{W}_{in}^l \cdot \mathbf{h}_s^{l-1}).
\end{equation}
Here, we also utilize $N$ random prefixes generated in the same manner as for the computing $\mathbf{v}_*$  \cite{meng2022locating}.

After obtaining the optimized $ \mathbf{{v}}_*$ and $\mathbf{k}_*$, we substitute them into Equation (\ref{solution}) to get the updated parameter $\mathbf{\hat{W}}$.

\section{Experiments}
In this section, we evaluate our \themodel by applying it to two datasets and assessing its performance on two auto-regressive LLMs. We aim to answer the following questions through experiments.
\begin{itemize}
\item {\bf Q1}: How does \themodel perform in multi-hop and single-hop reasoning evaluation compared with state-of-the-art editing methods ?
\item {\bf Q2}: How does the degree of erasure of old knowledge affect model's performance in multi-hop reasoning?
\item {\bf Q3}: What impact does our \themodel have on the retention of old knowledge?
\end{itemize}
\begin{table*}[th!]
\centering
\resizebox{\linewidth}{!}{%
\begin{tabular}{c|c|ccc|c|ccc}
\toprule
\multirow{2}{*}{\textbf{Editor}} &\multicolumn{4}{c}{\textbf{Correct Answer} \textcolor{dgreen}{\textbf{$\uparrow$}}} & \multicolumn{4}{c}{\textbf{Original Answer}\textcolor{red}{$\downarrow$}} \\ 
\cmidrule(lr){2-5} \cmidrule(lr){6-9} 
& Average Accuracy & Zero-Shot & Few-Shot & CoT & Average Accuracy & Zero-Shot & Few-Shot & CoT \\
\midrule
GPT-2 XL  & 7.12  & 6.32 & 8.66 & 6.37 & 33.14 & 19.12 & 41.50 & 38.80 \\
\midrule
FT & 4.45  & 2.70  & 4.58 & 11.68 & 34.92 & 26.07 & 23.12 & 55.56 \\
ROME & 10.32 & 7.52 & 11.68 & 11.76 & 20.40 & 12.66 & 24.92 & 23.61 \\
MEMIT  & 7.38 & {4.25} & 9.40 & 8.50 & 27.64 & 15.93 & 33.82 & 33.17 \\
\rowcolor{gray!20} \textbf{KELE} & \textbf{12.04} & \textbf{7.60}  & \textbf{14.05}   & \textbf{14.46} & \textbf{17.24 }  & \textbf{11.03} & \textbf{21.40} & \textbf{19.28} \\
\rowcolor{gray!20} $\Delta Improve$ & {16.67\%}  & {1.06\%} & {20.29\%}  & {22.96\%} &15.49\% &12.88\% &7.44\%&18.34\%\\
\midrule
GPT-J & 5.47 & 2.91  & 4.58 & 8.92 & 35.22 & 28.16 & 22.01 & 55.48 \\
\midrule
FT & 6.94 & 3.79  & 5.55  & 11.47 & 33.27 & 26.07 & 20.34 & 53.40 \\
ROME & 14.56 & 7.54 & 8.69 & 27.46 & 18.40 & 12.85 & 13.64 & 28.71  \\
MEMIT & 9.09 & 3.74 & 5.46  & 18.07 & 27.35 & 19.69 & 19.42 & 42.95 \\
\rowcolor{gray!20} \textbf{KELE} & 
\textbf{24.04} & \textbf{13.55} &
\textbf{15.40} & \textbf{43.18} &  \textbf{13.00} & \textbf{9.20} & \textbf{10.35} & \textbf{19.46} \\
\rowcolor{gray!20} $\Delta Improve$ & {65.11\%}  & {79.70\%}  & {77.21\%} & {57.25\%} &29.35\%&28.40\%&24.12\%&32.22\% \\

\bottomrule
\end{tabular}%
}
\caption{Performance comparison of editors on multi-hop questions of \textsc{MQuAKE-3K} dataset in terms of Multi-hop Accuracy (\%).\textcolor{dgreen}{\textbf{$\uparrow$}} indicates that higher values correspond to better performance, while \textcolor{red}{$\downarrow$} indicates that lower values correspond to better performance.}
\label{mquake_performance}
\end{table*}
\subsection{Experimental Setups}
\label{exper}
\subsubsection{Datasets and Evaluation Metrics}
We evaluate our \themodel on two representative datasets:\textbf{\textsc{MQuAKE-3K}} \cite{zhong2023maquake} and \textbf{\textsc{CounterFact}} \cite{meng2022locating}. Detailed descriptions of the datasets and evaluation metrics are provided in Appendix \ref{dataset} and \ref{metrics}.

\textbf{\textsc{MQuAKE-3K}} is a challenging dataset designed to assess models’ ability to perform multi-hop reasoning using newly edited knowledge. Each entry in this dataset involves multiple single-hop edits and includes multi-hop reasoning questions. This imposes stricter demands on the capability of edited LLMs to utilize the updated knowledge. Following \cite{zhong2023maquake}, we use \emph{Multi-hop Accuracy} to measure the performance of edited LLMs. To fully leverage the LLM’s reasoning ability, we employ three approaches when generating answers: Zero-shot, Few-shot, and Chain-of-Thought (CoT). The details of prompting are shown in Appendix \ref{prompt}. 

\textbf{\textsc{CounterFact}} is a dataset focused on evaluating LLMs’ ability to recall edited knowledge in a single-hop setting, as well as to assess the impact of editing operations on unrelated knowledge within the LLMs. Following \cite{meng2022locating}, we employ three widely used metrics for this dataset: \emph{Efficacy Score}, which measures the success rate of edits; \emph{Paraphrase Score}, which evaluates the model’s ability to accurately recall edited knowledge in paraphrased forms, testing its generalization ability; and \emph{Neighborhood Score}, which assesses whether irrelevant knowledge in the LLM is disturbed.

\subsubsection{Baselines}
Our experiments are conducted on GPT-2 XL (1.5B) \cite{radford2019language} and GPT-J (6B) \cite{gpt-j}, and we compare \themodel with state-of-the-art editing methods: Constrained Fine-Tuning (FT) \cite{zhu2020modifying}, ROME \cite{meng2022locating}, and MEMIT \cite{meng2022mass}. Implementation details for both baselines and \themodel are provided in Appendix \ref{baseline} and \ref{imp}.

\subsection{Performance Comparison (RQ1)}
\begin{figure}
  \centering
  \includegraphics[width=1\linewidth]{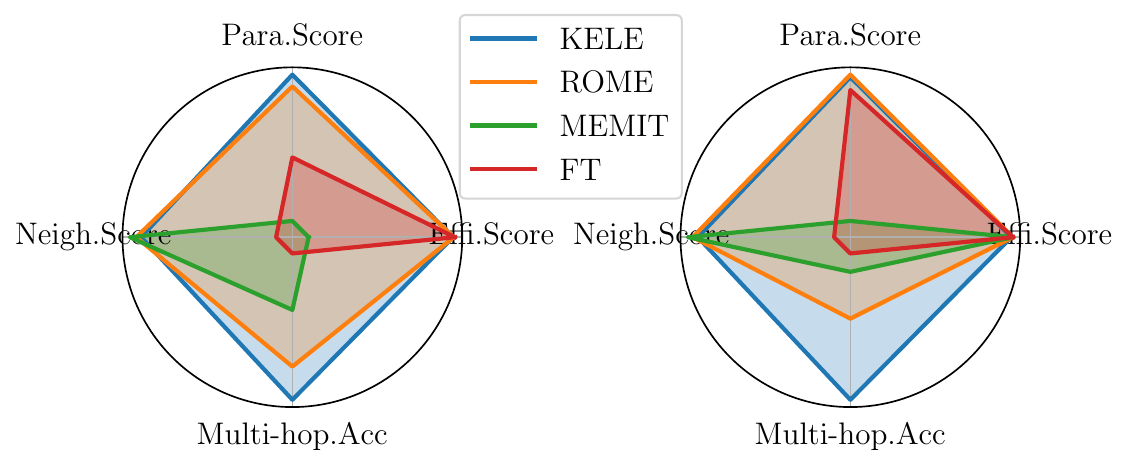}
  \caption{Comparative performance on GPT-2 XL and GPT-J across different metrics.}
  \label{map}
 \end{figure}

The performance of all editors on the \textsc{MQuAKE-3K} and \textsc{CounterFact} is presented in Tables \ref{mquake_performance} and \ref{cp}. Figure \ref{map} provides a comprehensive comparison of all editing methods across four metrics on both datasets, demonstrating that \themodel exhibits relatively balanced and superior performance across all metrics, particularly excelling in Multi-hop Accuracy, where it significantly outperforms other methods.

\subsubsection{\bf Results on \textsc{MQuAKE-3K}}
As shown in Table \ref{mquake_performance}, our \themodel outperforms all baselines by a significant margin across all evaluation metrics and settings. Specifically, \themodel demonstrates improvements of $16.67$ \% and $65.11$ \% in average multi-hop accuracy over the best baseline models for GPT-2 XL and GPT-J, respectively. This indicates that \themodel effectively enhances the ability of edited LLM in multi-hop reasoning tasks. Additionally, the multi-hop accuracy of \themodel in generating original answers decreased by an average of $15.49$\% and $29.35$ \% on GPT-2 XL and GPT-J, respectively, compared to the strongest baseline model. This suggests that the knowledge erasure operations in \themodel successfully mitigate the recall of old knowledge in the edited LLMs when performing complex reasoning tasks. These findings further support our hypothesis that residual old knowledge in the edited models is easily recalled during multi-hop reasoning. This recall causes the model to produce original answers to multi-hop questions, thereby weakening the LLM’s performance on such tasks

\subsubsection{\bf Results on \textsc{CounterFact}} 
From Table \ref{cp}, we find that our \themodel also achieves competitive results compared to all baselines. Unlike the \textsc{MQuAKE-3K} dataset, the \textsc{CounterFact} dataset focuses on evaluating the single-hop reasoning ability of the edited knowledge. \themodel achieves the best or near-best results in both Efficacy Score and Paraphrase Score on GPT-2 XL and GPT-J. This means that the operation of old knowledge erasure does not hinder the recall of edited knowledge. Although MEMIT achieves the best performance on Neighborhood Score, it performed poorly on Efficacy Score and Paraphrase Score.We attribute this to insufficient editing, resulting in the new knowledge not being effectively integrated into the LLMs. Compared to the ROME method, \themodel achieves a Neighborhood Score close to that of ROME, but significantly outperforms it in multi-hop reasoning. This suggests that our method integrates new knowledge with minimal disruption to the overall model, and that the erasure of old knowledge allows it to excel in complex multi-hop reasoning tasks.

\begin{table}[t!]
\centering
\resizebox{\linewidth}{!}{
\begin{tabular}{cc|ccc}
\toprule
 &\textbf{Editor} & \textbf{Effi.Score} & \textbf{Para.Score} & \textbf{Neigh.Score}  \\
\midrule
&{GPT-2 XL}  & 22.20 & 24.70 & 78.10  \\
\midrule
& FT      & \textbf{100.00} & 87.90 & \textcolor{gray}{40.40}  \\
& ROME    & \underline{99.95}  & \underline{96.48} & \underline{75.44}  \\
& MEMIT   & 93.79  & 80.22 & \textbf{77.05}  \\
\rowcolor{gray!20}& \themodel & 99.92  & \textbf{97.90} & 74.21 \\

\midrule
&{GPT-J } & 16.30 & 18.60 & 83.00  \\
\midrule
& FT      & 100.00 & 98.80 & \textcolor{gray}{10.30}  \\
& ROME    & 100.00 & \textbf{99.22} & \underline{78.89}  \\
& MEMIT   & 100.00 & 95.23 & \textbf{81.26} \\
\rowcolor{gray!20}& \themodel & \underline{99.90} & \underline{99.15} & {76.39} \\
\bottomrule
\end{tabular}}
\caption{Performance comparison on \textsc{Counterfact} in terms of Efficacy Score (\%), Paraphrase Score (\%), and Neighborhood Score (\%). The best performance is highlighted in boldface, and the second-best is underlined. Gray numbers indicate a clear failure on the metric. }
\label{cp}
\end{table}

\subsection{Impact of Erasure Intensity (RQ2)}
\begin{figure}[t!]
	\centering
	\subfloat[Original Answer]{\includegraphics[scale=0.62]{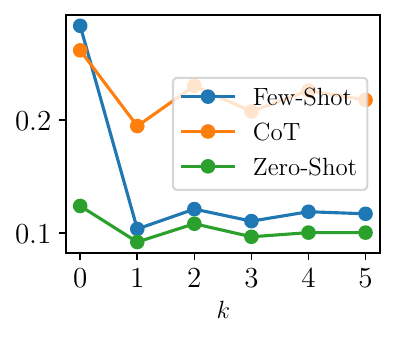}
 \label{gptj-k-o}
	}
	\subfloat[Correct Answer]{\includegraphics[scale=0.62]{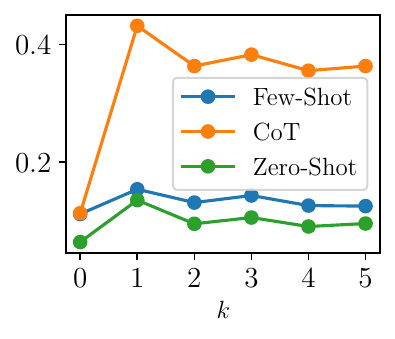}
  \label{gptj-k-n}
	}
	\caption{Performance of edited GPT-J with different $k$.
	}
	\label{gptj-k}
\end{figure}

The hyperparameter $k$ of Equation (\ref{eq-e}) represents the degree of erasure of old knowledge. A larger $k$ indicates a higher degree of erasure, and vice versa. To investigate the impact of varying erasure intensities on the model, we conduct experiments with various $k$ values on the \textsc{MQuAKE-3K} dataset. The results, shown in Figure \ref{gptj-k}, lead to the following observations: As $k$ increases, the erasure of old knowledge is enhanced, and the accuracy of generating original answers for multi-hop questions gradually decreases. This further validates that residual old knowledge after editing encourages models to revert to original answers in multi-hop questions. Furthermore, the edited GPT-J achieve its best performance at $k=1$, with the highest accuracy in generating correct answers. Beyond this point, as $k$ continues to increase, the performance of the models either stabilizes or declines. This may be due to excessively high erasure intensity. While it reduces the likelihood of generating original answers, it may also introduce other disruptions to the model, ultimately weakening its reasoning ability.

\begin{figure}
  \centering
  \includegraphics[width=1\linewidth]{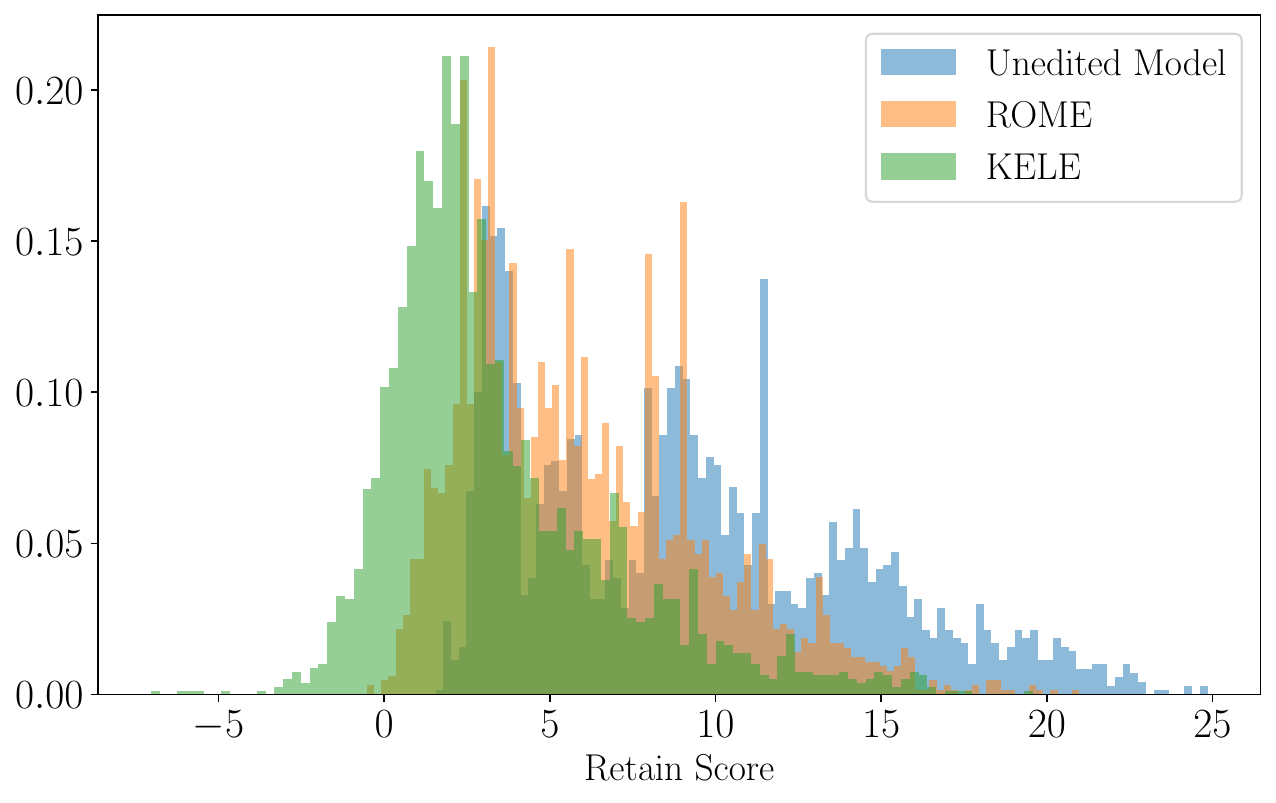}
  \caption{The distribution of Retain Score.}
  \label{distribution}
 \end{figure}

\subsection{The impact on Old Knowledge (RQ3)}

To investigate the impact of \themodel on old knowledge $(s,r,o)$, we examine the distribution of Retain Score in three models: the unedited LLM (GPT-J), the LLM edited with ROME, and the LLM edited with \themodel. The experimental results are presented in Figure \ref{distribution}. From the results, we observe that the unedited model exhibits the highest Retain Scores, with a significant density around $10$ to $15$, indicating substantial retention of old knowledge. The ROME-edited model shows a reduction in Retain Score, shifting the distribution leftward, but still retains a noticeable amount of old knowledge, particularly in the $5$ to $10$ range. In contrast, the KELE demonstrates the most significant reduction, with a peak near lower Retain Scores. These results demonstrate that \themodel effectively erases residual old knowledge, which is crucial for enhancing the model’s performance in multi-hop reasoning tasks.

\section{Conclusion}
In this paper, drawing inspiration from neuroscience, we propose and experimentally validate that the poor performance of current editing methods in multi-hop scenarios is due to the retention of single-hop old knowledge. This residual knowledge causes edited LLMs to revert to original answers when responding to multi-hop questions. Building on this foundation, we introduce a  simple yet effective knowledge editing method that incorporates a knowledge erasure mechanism. This method combines an old knowledge erasure strategy with a rank-one model editing framework, eliminating old knowledge while injecting new knowledge. Experimental results on two LLMs, supported by extensive analysis, demonstrate the effectiveness and superiority of \themodel in multi-hop reasoning tasks.

\bibliography{aaai25}

\newpage
\appendix

\section{Dataset}
\label{dataset}
We evaluate our \themodel on two representative datasets:\textbf{\textsc{MQuAKE-3K}} \cite{zhong2023maquake} and \textbf{\textsc{CounterFact}} \cite{meng2022locating}.
\subsection{Details of \textsc{MQuAKE-3K} Dataset}
{\textsc{MQuAKE-3K}} is a challenging dataset designed to assess models’ ability to perform multi-hop reasoning using newly edited knowledge. Each entry in this dataset involve multiple edits and includes multi-hop reasoning questions that require reasoning from $2$ to $4$ hops to answer correctly. This imposes stricter demands on the capability of edited LLMs to utilize the updated knowledge. 
Table \ref{mquake_sample} provides an example from \textsc{MQuAKE-3K} dataset. In this example, two edits are required: inserting the knowledge \emph{(Lou Pearlman, is a citizen of, India)} and \emph{(India, The capital of, Taloga)}. Accordingly, a 3-hop question ``\emph{What is the capital of the country to which Lou Pearlman belonged?}'' is constructed to assess the post-edit models's ability to ulitze edited knowledge and its related information.  Following \cite{zhong2023maquake}, our evaluation focuses on a subset of $3000$ entries, evenly distributed across $\{2,3,4\}$-hop questions, with each category comprising $1000$ entries. 

\subsection{Details of \textsc{CounterFact} Dataset}
Table \ref{cf_sample} presents an example from the \textsc{CounterFact} dataset. Each entry includes an edit request, several paraphrase prompts, and neighborhood prompts. In this example, the edit request aims to change the model's knowledge of \emph{The mother tongue of Go Hyeon-jeong} from \emph{Korean} to \emph{French}. Paraphrase prompts are semantic variations of the target prompt, while neighborhood prompts involve the same relation but with a different subject, whose knowledge should remain unaffected by the edit.

\begin{table*}[h!]
  \centering
  \resizebox{\linewidth}{!}
  {%
  \begin{tabularx}{\textwidth}{lX} 
    \toprule
    \textbf{Property} & \textbf{Value} \\
    \midrule
    Edit Request 1 & \{Lou Pearlman \} is a citizen of \textit{United States of America $\to$ India} \\
    Edit Request 2 & The capital of \{India\} is \textit{New Delhi $\to$ Taloga} \\
    New Question & What is the capital of the country to which Lou Pearlman belonged? \\
    Original Relation & (Lou Pearlman, a citizen of, United States of America), (United States of America, the capital of, Washington) \\
    Original Answer & Washington \\
    New Relation & (Lou Pearlman,  a citizen of, India), (India, the capital of, Taloga)\\
    New Answer & Taloga\\
    \bottomrule
  \end{tabularx}
  }
  \caption{An Example of \textsc{MQuAKE} dataset}
  \label{mquake_sample}
\end{table*}

\begin{table*}[t]
  \centering
  \resizebox{\linewidth}{!}
  {%
  \begin{tabularx}{\linewidth}{lX} 
    \toprule
    \textbf{Property} & \textbf{Value} \\
    \midrule
    Edit Request & The mother tongue of \{Go Hyeon-jeong\} is \textit{Korean $\to$ French}  \\
    Efficacy\_prompt & The mother tongue of Go Hyeon-jeong is \\
    Paraphrase\_prompt & It won the Governor General's Literary Award the same year. Go Hyeon-jeong spoke the language \\
    Neighborhood\_prompt & The native language of Gong Ji-young is \\
    \bottomrule
  \end{tabularx}
  }
  \caption{An Example of \textsc{CounterFact} dataset}
  \label{cf_sample}
\end{table*}

\section{Evaluation Metrics}
\label{metrics}
For each instance $d=(\mathcal{E},\mathcal{Q}, a,a^*)$ in the \textsc{MQuAKE} dataset, the multi-hop accuracy after editing is defined as:
 \begin{equation*}
    \frac{1}{|\mathcal{Q}|}\sum_{q \in \mathcal{Q}} \mathbb{I}\left[\mathcal{F}{'}(q) = a^{*})\right].
    \end{equation*}
We report the averaged multi-hop accuracy in our evaluation.

For the \textsc{CounterFact} dataset, we use three widely-used metrics \cite{meng2022locating, meng2022mass}, {Efficacy Score}, {Paraphrase Score}, and {Neighborhood Score} to evaluate all editors. Each metric is calculated as follows:

\textbf{Efficacy Score} is to test whether the post-edit LLMs can correctly recall the new target entity when given the edit prompt $p(s, r)$. It is calculated by
    \begin{equation*}
        \mathbb{E}\left[\mathbb{\mathbb { I }}\left[\mathrm{P}_{\mathcal{F}'}\left(o^{*}\mid p(s,r)\right)>\mathrm{P}_{\mathcal{F}'}\left(o\mid p(s,r)\right)\right]\right].
    \end{equation*}

\textbf{Paraphrase Score} measures the performance of the post-edit LLM on rephase prompt set ${P}^P$ of edit prompt $p(s,r)$.  The calculation is similar to the Efficacy Score:
    \begin{equation*}
    \mathbb{E}_{p \in {P}^P}\left[\mathbb{\mathbb { I }}\left[\mathrm{P}_{\mathcal{F}'}\left(o^{*}\mid p \right)>\mathrm{P}_{\mathcal{F}'}\left(o\mid p \right)\right]\right].
    \end{equation*}

\textbf{Neighborhood Score} measures whether the post-edit LLM assigns the higher probability to the correct fact on the prompt set ${P}^N$, which consists of distinct but semantically similar prompts $p(s,r)$. The calculation is defined as:
    \begin{equation*}
    \mathbb{E}_{p \in {P}^N}\left[\mathbb{\mathbb { I }}\left[\mathrm{P}_{\mathcal{F}'}\left(o^{*}\mid p \right)<\mathrm{P}_{\mathcal{F}'}\left(o\mid p \right)\right]\right].
    \end{equation*}
This metric can assess the extent of the impact that edits have on unrelated knowledge.

\section{Baselines}
\label{baseline}
Our experiments are conducted on GPT-2 XL (1.5B) \cite{radford2019language} and GPT-J (6B) \cite{gpt-j}, and we compare \themodel with the following state-of-the-art editing methods:

{\bf Constrained Fine-Tuning (FT)} \cite{zhu2020modifying} involves fine-tuning specific layers of the LLM's parameters directly using gradient descent, while imposing a norm constraint on the weight changes to prevent catastrophic forgetting.

{\bf ROME} \cite{meng2022locating} is based on the hypothesis that knowledge in LLMs is stored in the FFN module, and uses optimization to update a FFN layer to insert knowledge.

{\bf MEMIT}  \cite{meng2022mass} builds on the ROME method, specializing in batch-editing tasks by performing edits on a range of FFN layers.

\section{Implementation Details}
\label{imp}
We implement our \themodel method using \textbf{PyTorch}\footnote{\url{https://pytorch.org/}}. 
For the other baselines, we conduct our experiments using the code provided by ROME \cite{meng2022locating}, ensuring that all settings, including hyperparameters, are consistent with \cite{meng2022locating,meng2022mass}. For our KELE, on the \textsc{MQuAKE-CF-3K} dataset, editing operation is performed at layer $12$ for GPT-J with the optimal \(k\) value of 1, selected after searching within \(k = \{0, 1, 3, 4, 5\}\). For GPT-2 XL, editing is carried out at layer $24$, and the optimal \(k\) value of $4$ chosen from the same search space. On the \textsc{CounterFact} dataset, editing is performed at layer $5$ for GPT-J with \(k = 5\) and at layer $17$ for GPT-2 XL with \(k = 5\). Other parameters are kept consistent with those used in ROME. We run
the evaluation five times with different random seeds and report the mean value of each method. Our experiments are conducted on NVIDIA Tesla A100 (80G) and AMD EPYC 7742 CPU.

\section{Prompt used in \textsc{MQuAKE}}
\label{prompt}
\begin{figure*}[th!]
    \centering
    \includegraphics[width=1.0\linewidth]{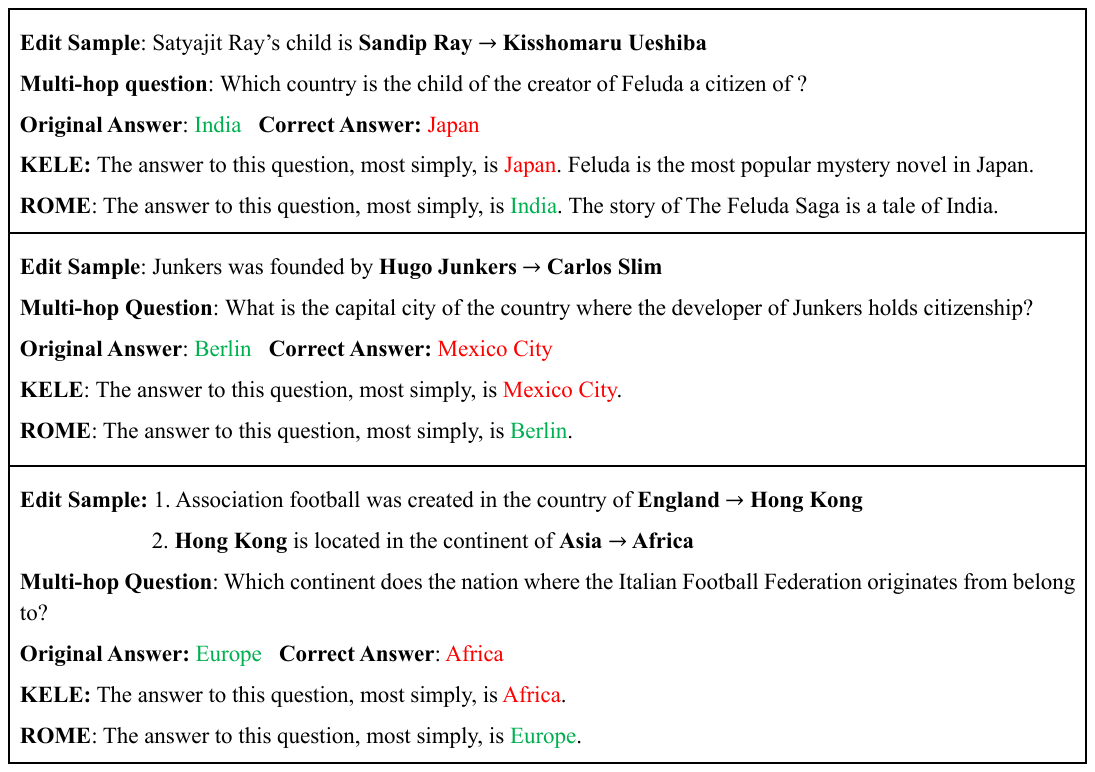}
\caption{The template of the few-shot prompt.}
    \label{few-shot}
\end{figure*}

\begin{figure*}[th!]
 \centering
    \includegraphics[width=1.0\linewidth]{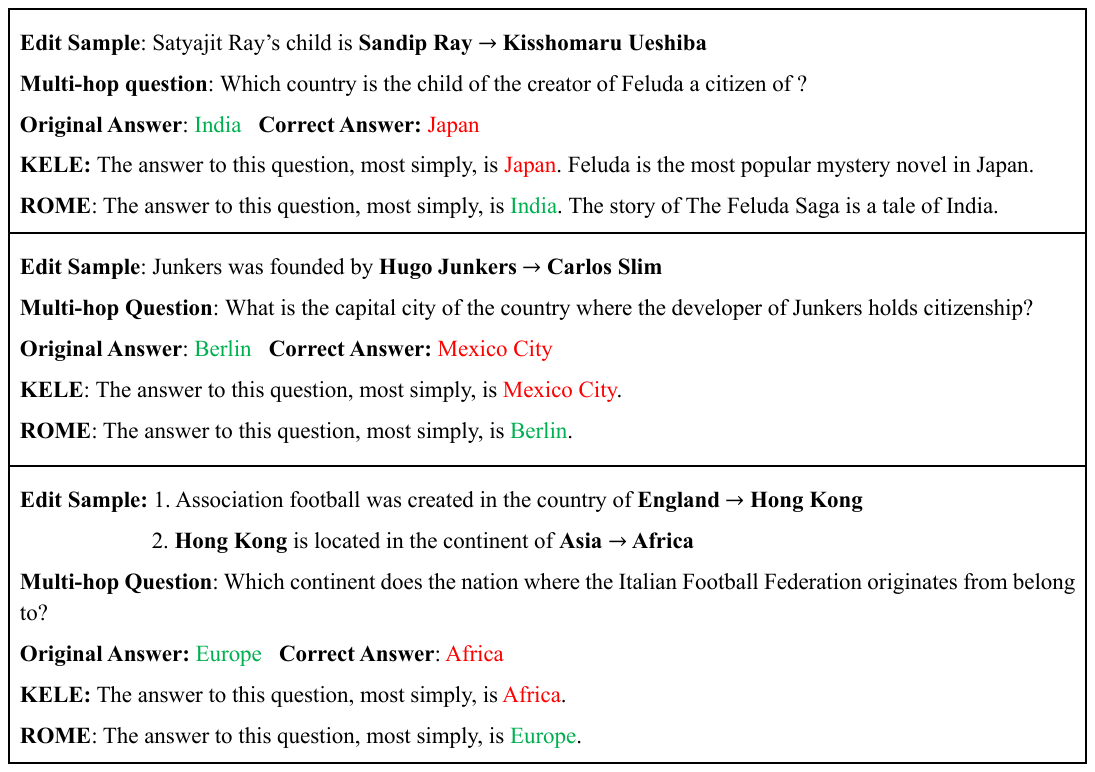}
\caption{The template of the chain-of-shot prompt.}
    \label{cot}
\end{figure*}

To fully leverage the LLM’s reasoning ability, we employ three approaches when generating answers: Zero-shot, Few-shot, and Chain-of-Thought (CoT). The templates of few-shot prompt and CoT prompt are shown in Figures \ref{few-shot} and \ref{cot}. 

\begin{figure}[h]
	\centering
	\subfloat[Original Answer]{\includegraphics[scale=0.62]{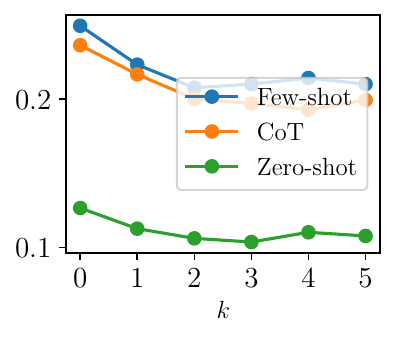}
 \label{gpt2-k-o}
	}
	\subfloat[Correct Answer]{\includegraphics[scale=0.62]{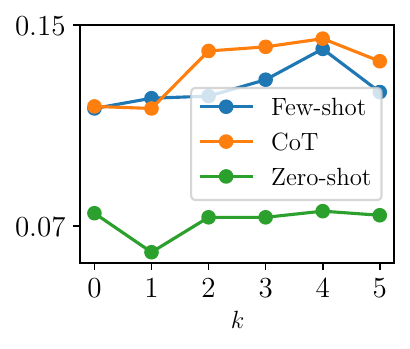}
  \label{gpt2-k-n}
	}
	\caption{Performance of edited GPT-2 XL with different $k$.
	}
	\label{gpt2-k}
\end{figure}

\section{Case Study}
\begin{figure*}[th!]
    \centering
    \includegraphics[width=1.0\linewidth]{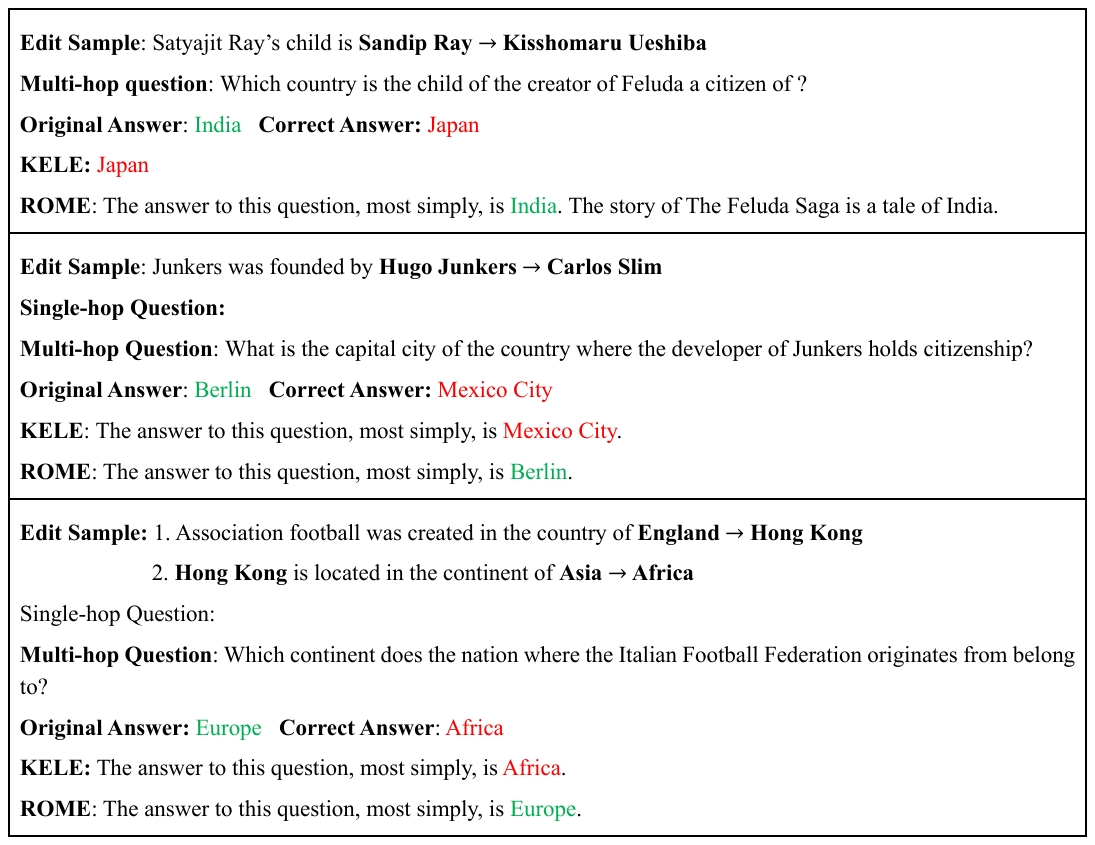}
\caption{Case 1. GPT-J generation examples of \themodel and ROME. \textcolor{dgreen}{\textbf{Green}} indicates the correct answers to single-hop and multi-hop questions, while \textcolor{red}{\textbf{Red}} indicates the original answers.}
    \label{case1}
\end{figure*}

\begin{figure*}[th!]
    \centering
    \includegraphics[width=1.0\linewidth]{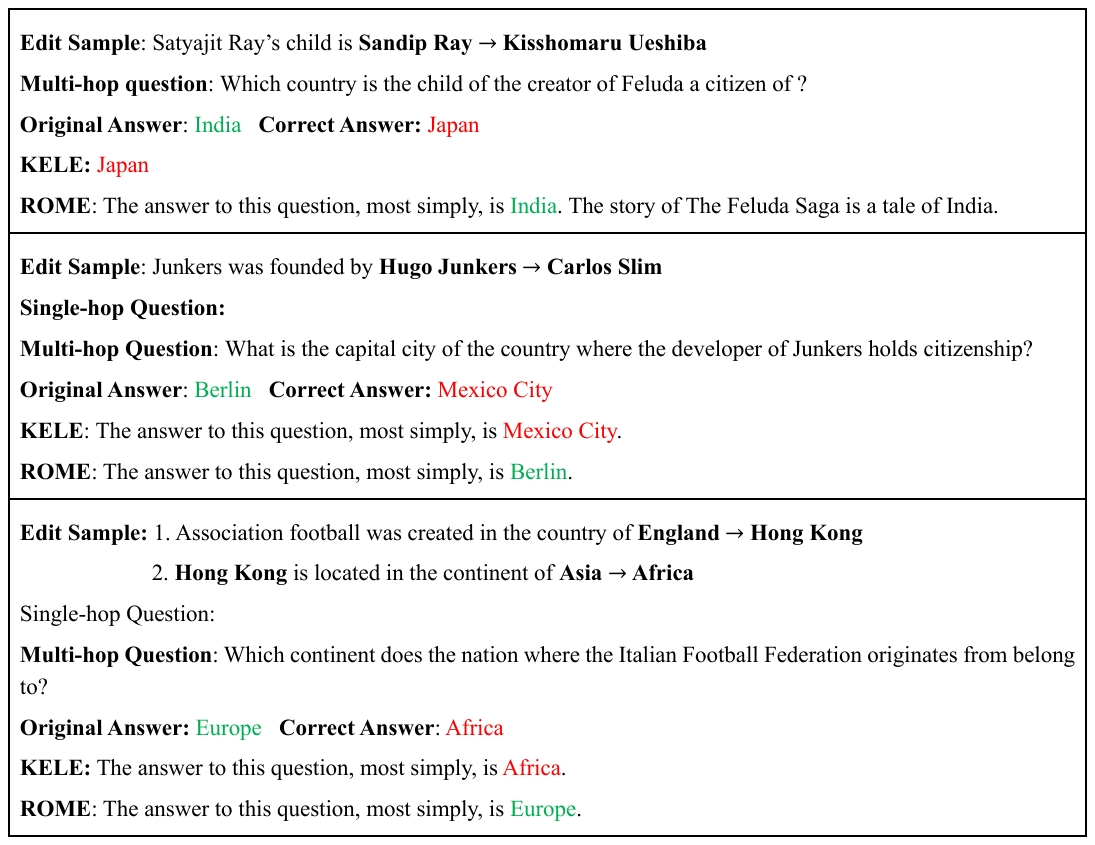}
\caption{Case 2. GPT-J generation examples of \themodel and ROME. \textcolor{dgreen}{\textbf{Green}} indicates the correct answers to single-hop and multi-hop questions questions, while \textcolor{red}{\textbf{Red}} indicates the original answers.}
    \label{case2}
\end{figure*}

\begin{figure*}[th!]
    \centering
    \includegraphics[width=1.0\linewidth]{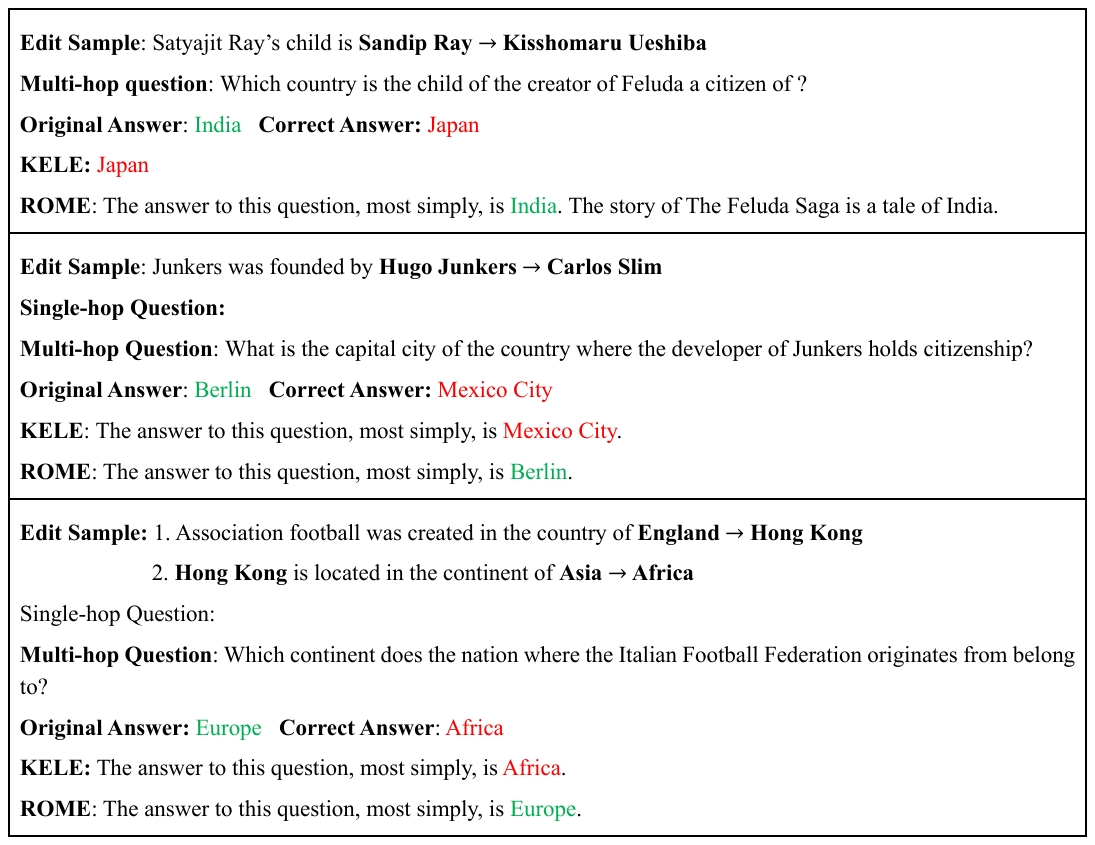}
\caption{Case 3. GPT-J generation examples of \themodel and ROME. \textcolor{dgreen}{\textbf{Green}} indicates the correct answers to single-hop and multi-hop questions, while \textcolor{red}{\textbf{Red}} indicates the original answers.}
    \label{case3}
\end{figure*}

In this section, we present several generation examples on GPT-J using two knowledge editing methods: \themodel and ROME, to demonstrate the efficacy of \themodel in enhancing multi-hop reasoning. The generation examples are  illustrated in Figures \ref{case1}, \ref{case2}, and \ref{case3}.

In the first two cases (Figures \ref{case1} and \ref{case2}), a single piece of knowledge is edited, such as changing “Satyajit Ray’s child is Sandip Ray” to “Satyajit Ray’s child is Kisshomaru Ueshiba.” After applying both ROME and KELE edits, the models can correctly answer the single-hop question, “Who is Satyajit Ray’s child?” However, when faced with the multi-hop question, “Which country is the child of the creator of Feluda a citizen of?”, the ROME-edited model still generates the original answer, “India.” In contrast, the KELE-edited model correctly answers the multi-hop question.

Figure \ref{case3} (Case 3) presents a more complex scenario involving the editing of two single-hop facts. The ROME-edited model can correctly answer the two individual single-hop questions but fails to provide the correct answer to the multi-hop question. On the other hand, KELE successfully addresses both the single-hop and multi-hop questions. These findings further validate that \themodel enhances the reasoning capabilities of the edited model in multi-hop tasks by effectively eliminating residual old single-hop knowledge.

\section{Impact of Erasure Intensity}
Figure \ref{gpt2-k} shows the performance of GPT-2 XL with various $k$ values on the \textsc{MQuAKE-3K} dataset.  The results indicate that as as $k$ increases, the erasure of old knowledge is enhanced, and the accuracy of generating original answers for multi-hop questions gradually decreases. This further validates that residual old knowledge after editing encourages models to revert to original answers in multi-hop questions. Furthermore, the edited GPT-2 XL achieve its best performance at $k=4$, with the highest accuracy in generating correct answers. Beyond this point, as $k$ continues to increase, the performance of the models either stabilizes or declines. This may be due to excessively high erasure intensity. While it reduces the likelihood of generating original answers, it may also introduce other disruptions to the model, ultimately weakening its reasoning ability.

\end{document}